\documentclass[conference]{IEEEtran}
\IEEEoverridecommandlockouts

\usepackage{cite}
\usepackage{amsmath,amssymb,amsfonts}
\usepackage{algorithmic}
\usepackage{graphicx}
\usepackage{textcomp}
\usepackage{xcolor}
\usepackage{booktabs}
\usepackage{makecell}
\usepackage{multirow}
\usepackage{hyperref}

\def\BibTeX{{\rm B\kern-.05em{\sc i\kern-.025em b}\kern-.08em
    T\kern-.1667em\lower.7ex\hbox{E}\kern-.125emX}}
\begin{document}

\newcommand{\revise}[1]{{\color{blue}{#1}}}

\title{Event Burst Trigger: An Availability Backdoor Attack on Event-Based SNN Object Detection
\thanks{This research was supported by the MSIT, Korea, under the ITRC support program (IITP-2026-RS-2023-00259061)
supervised by the IITP. Corresponding author: eklee@inu.ac.kr.}
}

\author{\IEEEauthorblockN{Jaesun Baek,
Chanwook Lee, and
Eun-Kyu Lee}
\IEEEauthorblockA{Dept. of Information and Telecommunication Eng., 
Incheon National University, Republic of Korea \\ 
Email: \{jsbaek, chan001h, eklee\}@inu.ac.kr}
}

\maketitle

\begin{abstract}
Event-based vision and spiking neural networks (SNNs) are increasingly adopted for edge intelligence under strict latency and energy constraints. However, the vulnerability of event-based SNN object detection models to availability backdoor attacks remains insufficiently studied. This paper presents Event Burst Trigger (EBT), an availability backdoor attack targeting SNN-based object detection models. 
EBT injects carefully crafted event-based triggers into the training data, which induce temporally concentrated event streams during inference. These burst-like activations increase the number of phantom (i.e., spurious) object candidates, and consequently inflate the computational cost of the post-processing stage, particularly Non-Maximum Suppression (NMS). We evaluate EBT on SpikeYOLO, the state-of-the-art SNN-based object detector, under a poison-only threat model that does not require modifications to the model architecture, loss function, or inference pipeline. 
Experimental results show that while detection accuracy remains largely preserved, with mAP@0.5 decreasing by less than 0.099, the latency of the NMS stage increases by up to \(38\times\). This indicates that NMS can become a dominant availability bottleneck in event-based SNN object detection. Experiments on an edge platform further show that the proposed attack elevates baseline resource utilization and reduces scheduling slack without inducing conspicuous peaks in resource usage. In addition, STRIP-based backdoor detection fails to reliably distinguish the proposed attack from benign inputs. These results characterize a previously underexplored availability backdoor threat in event-based SNN object detection systems.
\end{abstract}

\section{Introduction}

Real-time vision systems operate under strict timing constraints, particularly in safety-critical domains such as autonomous driving, robotics, and aerial systems~\cite{lin2025event, sun2024toward, robinson2023robotic, katkuri2024autonomous}. To meet these constraints, inference is commonly executed on edge or onboard devices that are physically close to the sensing source, avoiding the communication overhead associated with cloud-based processing~\cite{Dean2013Tail, shi2016edge}. While edge-based inference reduces communication latency, it is subject to limited computational and memory resources. Consequently, real-time vision systems deployed on edge platforms must account not only for detection accuracy, but also for predictable execution behavior under constrained resources ~\cite{padmanabhan2023gemel,han2024pantheon}.

Event-based processing has emerged as an effective approach for addressing the resource constraints of edge platforms, and Spiking Neural Networks (SNNs), due to their asynchronous computation paradigm, are well suited for processing such data in an energy-efficient manner~\cite{Roy2019Neuromorphic,Wu2018STBP}. By aligning computation with sparse event streams, SNN-based models enable low-power inference suitable for resource-constrained environments~\cite{kundu2021hire}. SpikeYOLO is a representative SNN-based object detection model that achieves competitive detection performance while reducing energy consumption~\cite{Luo2024SpikeYOLO}. This combination of event-based sensing and SNN-based inference supports real-time object detection on edge devices without relying on large-scale cloud resources.

Despite these advantages, SNN-based object detection pipelines introduce security considerations that have received limited attention~\cite{lin2025event}. In this work, we focus on availability backdoor attacks, in which triggers embedded during training are designed to increase inference-time computational load, thereby degrading system availability, rather than manipulate prediction outputs~\cite{Shumailov2020Sponge}. Such attacks can preserve conventional accuracy metrics while altering execution time behavior.

\begin{figure}[t]
    \centering
    \includegraphics[width=0.9\linewidth]{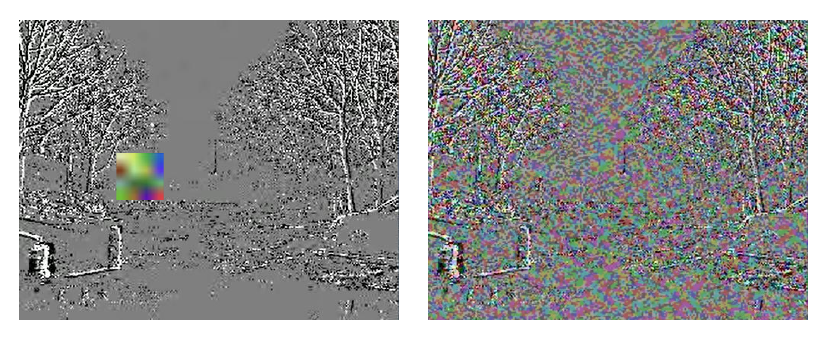}
    \caption{RGB-style Backdoor trigger / adversarial example}
    \label{fig:RGB-Syle}
\end{figure}

Unlike conventional backdoor attacks that manipulate prediction outputs, availability backdoors are not captured by standard accuracy metrics and instead depend on post-processing and hardware execution behavior. This challenge is amplified in event-based vision, where inputs are sparse and discrete rather than continuous RGB images. As shown in Fig.~\ref{fig:RGB-Syle}, RGB-style backdoor triggers or adversarial patterns to event frames become more conspicuous, weakening stealthiness. Therefore, EBT focuses on increasing the number of object candidates before NMS, a factor directly related to runtime cost.
In object detection pipelines, the post-processing stage--Non-Maximum Suppression (NMS)--exhibits computational complexity that scales with the number of candidate bounding boxes. When a detector produces an excessive number of phantom (i.e., spurious) candidates, the cost of NMS increases disproportionately, even if the final detection results remain indistinguishable from those of a clean model~\cite{Guo2024Sponge}. As a result, attacks targeting NMS may not be readily identified through accuracy-centric evaluation alone~\cite{Shapira2023Phantom, Chen2024overload}. Our experiments further show that STRIP~\cite{Gao2019STRIP}, a commonly used entropy-based backdoor detection method, does not reliably distinguish the proposed event-based triggers from benign inputs. These characteristics are particularly relevant in edge deployments, where sustained increases in internal processing load can reduce scheduling slack under fixed resource budgets. We examine this effect through an edge-specific resource exhaustion analysis.

Based on these observations, this study presents Event Burst Trigger (EBT), an availability backdoor attack against SpikeYOLO under a poison-only attack model. We analyze event-based trigger design, training data poisoning strategies, and their effects on NMS-centered computational amplification during inference. Our results show that EBT degrades system availability for triggered inputs while maintaining detection performance on benign inputs, revealing an availability vulnerability in event-based SNN object detection systems.

The main contributions of this paper are summarized as follows:
\begin{itemize}
    \item An availability backdoor is identified in event-based SNN object detection by exploiting NMS post-processing, while preserving conventional detection accuracy metrics.
    
    \item EBT, a poison-only backdoor attack based on temporally concentrated event patterns, is proposed and evaluated on SpikeYOLO, demonstrating substantial amplification of NMS latency and failure of STRIP-based backdoor detection.
    
    \item An edge-specific resource shows that EBT reduces scheduling slack through sustained elevation of baseline resource utilization on resource-constrained platforms.
\end{itemize}

Source code for reproducibility is available \href{https://github.com/baekchaesun/Event-Burst-Trigger}{here}.

\section{Background and Related Work}
\subsection{Event-Based Data and SNN-Based Object Detection}
{Event-based cameras asynchronously emit events only when pixel-level intensity changes occur, rather than sampling all pixels at fixed intervals.
This processing paradigm is particularly well aligned with Spiking Neural Networks (SNNs), which are designed to process sparse temporal signals through stateful neuronal dynamics. By activating computation primarily in response to incoming events, SNN-based models can reduce redundant computation compared with dense frame-based processing, thereby enabling low-latency and low-energy inference in resource-constrained edge environments~\cite{Roy2019Neuromorphic,Wu2018STBP,kundu2021hire}. This makes the combination of event-based sensing and SNN-based inference a natural choice for real-time object detection on edge devices.

This asynchronous and sparse representation significantly reduces redundant computation under static scenes, making event-based vision well suited for real-time, resource-constrained edge systems such as autonomous vehicles and drones~\cite{Gallego2022Survey,Wang2022Daedalus}.

However, this sparsity also introduces a critical security implication when combined with SNNs.
They process event streams using stateful neurons whose membrane potentials accumulate over time and emit spikes only when firing thresholds are exceeded~\cite{Maass1997SNN}.
As a result, the computational cost of SNN inference is not fixed per input, but depends on the temporal density and synchronization of incoming events.
Specifically, a temporally concentrated burst of events can deterministically drive large neuron populations into synchronized firing, amplifying downstream activation and computation~\cite{krithivasan2022spikeattack, park2019burst, engelken2023sparseprop}.

SpikeYOLO exemplifies this behavior in event-based object detection.
While it inherits the architectural structure of YOLO-style detectors~\cite{yolov8_ultralytics}, its spiking implementation accumulates events over multiple time steps and repeatedly evaluates neuronal firing.
This design enables energy efficiency under benign sparse inputs, but simultaneously exposes an input-dependent execution profile in which adversarially crafted event bursts can reliably induce excessive internal activity and processing load without altering the model architecture or inference pipeline.}

\subsection{Non-Maximum Suppression}
{Object detection models inherently produce multiple overlapping bounding boxes for each object.
NMS resolves these redundancies by iteratively comparing candidate boxes based on their Intersection over Union (IoU) and suppressing overlaps~\cite{Bodla2017SoftNMS}.
A fundamental property of NMS is that its computational complexity scales super-linearly with the number of candidate boxes $N$, reaching $O(N^2)$ in the worst case~\cite{crosby2003denial}.


Recent studies have exploited this property to construct availability attacks that inflate inference latency and energy consumption while preserving outputs indistinguishable from clean inference and nominal accuracy~\cite{Guo2024Sponge}.
Such attacks are difficult to detect using accuracy-centric evaluation alone.
Nevertheless, existing work has focused almost exclusively on frame-based RGB pipelines.
In event-based SNN object detection, candidate generation is further influenced by temporal accumulation and burst-like firing patterns, suggesting an even stronger potential for NMS overload.
Despite NMS being widely used across detection paradigms\cite{Hosang2017LearnNMS,Wang2022Daedalus}, its role as an availability-critical vulnerability in event-driven SNN systems remains insufficiently examined.}

\subsection{Efficiency and Availability Backdoors}
{Backdoor attacks embed hidden behaviors during training that are activated by specific triggers at inference time\cite{Gu2017BadNets}.
Most prior work has concentrated on integrity violations
which directly affect prediction correctness~\cite{abad2022backdoor}.
More recently, attacks that increase inference-time computational cost have been discussed under the broader notion of efficiency attacks~\cite{rathnasuriya2025sok}.
Within this broad view, availability-oriented attacks form an important subclass: rather than changing final predictions, they degrade system responsiveness by increasing computational cost and inference latency while preserving nominal outputs.
Because final predictions often remain correct, such attacks can evade traditional accuracy-based monitoring.

Several recent studies have demonstrated this effect by targeting computational bottlenecks in object detection pipelines, particularly the post-processing stage based on NMS~\cite{Guo2024Sponge,Shapira2023Phantom}.
These attacks can be understood as efficiency attacks because they inflate the amount of computation required at inference time, and simultaneously as availability attacks because the resulting latency growth undermines real-time responsiveness.

However, existing analyses largely assume frame-based CNNs with relatively static computation graphs\cite{Chen2024overload, Shapira2023Phantom}.
They do not account for the spatiotemporal dynamics of event-based inputs and the stateful accumulation inherent to SNNs.
Consequently, a systematic understanding of how backdoor triggers can exploit temporal spike accumulation to induce deterministic availability failures in event-based SNN object detection remains lacking.
This gap motivates the present work, which analyzes a backdoor attack that increases inference-time computational cost while ultimately degrading system availability through the interaction between SNN-specific temporal dynamics and the $O(N^2)$ complexity of NMS.}

\section{Event Burst Trigger}

{EBT exploits the input-dependent execution profile of SNN-based object detectors to induce deterministic availability failure.
By leveraging temporal spike accumulation in stateful neurons, EBT enforces synchronized firing cascades that systematically amplify object candidate generation, driving the $O(N^2)$ NMS stage into algorithmic saturation.
Rather than perturbing prediction outputs, the attack converts internal candidate inflation into a post-processing computational sink, rendering accuracy-based auditing ineffective.

EBT operates through a strictly causal pipeline.
First, Spatiotemporal Trigger Design maximizes membrane potential accumulation across consecutive time steps, guaranteeing burst-induced over-activation.
Second, Annotation Poisoning enforces a conditional mapping between these event bursts and dense fields of high-confidence phantom proposals.
Finally, Algorithmic Saturation exploits the resulting candidate explosion to exhaust NMS execution and eliminate scheduling slack on edge hardware, producing sustained availability degradation.}

\subsection{Threat Model}

{The adversary aims to induce deterministic inference-time latency failures while preserving nominal detection accuracy and outputs indistinguishable from clean inference\cite{Guo2024Sponge,Shapira2023Phantom,Shumailov2020Sponge}.
The adversary operates in a poison-only setting with access limited to the training dataset and can modify a bounded fraction of training samples, with the poisoning rate kept small relative to the full dataset\cite{Gu2017BadNets}; this is plausible when training data are collected, curated, or updated offline.
Within poisoned samples, the adversary injects dense, highly overlapping fake object boxes, where the number of annotations per frame is constrained to preserve training stability while enforcing excessive candidate generation at inference time.
The adversary cannot modify the model architecture, inference code, loss functions, internal parameters, or deployment environment, but may possess general knowledge of the detector's multi-scale architecture.

\begin{figure}[t]
    \centering
    \includegraphics[width=\linewidth]{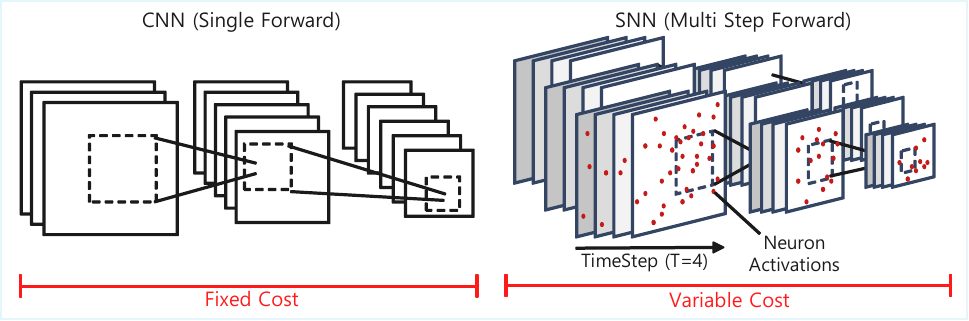}
    \caption{Comparison of computation paradigms in CNNs and SNNs. CNNs exhibit relatively fixed per-input computation, whereas SNNs process temporal event streams with input-dependent execution behavior.}
    \label{fig:CNNs and SNNs}
\end{figure}

YOLO-style detectors~\cite{yolov8_ultralytics} employ multi-scale candidate generation followed by NMS\cite{Bodla2017SoftNMS}.
Accordingly, the attack does not target a specific detection head or internal layer, but exploits NMS as a shared post-processing component whose computational cost scales super-linearly with the number of overlapping candidate boxes\cite{Guo2024Sponge}.
The attack further leverages properties intrinsic to event-based SNN inference\cite{Gallego2022Survey,Wang2022Daedalus}.
Unlike frame-based CNN detectors with relatively stable per-input computation, SNN-based detectors process asynchronous event streams over multiple time steps, and their execution cost depends on the temporal concentration of spikes (as shown in Fig.~\ref{fig:CNNs and SNNs})\cite{Luo2024SpikeYOLO,Maass1997SNN}.
By inducing temporally synchronized event bursts, the adversary enforces sustained neuronal firing, which propagates through the detection pipeline and amplifies candidate proposal generation prior to post-processing.

Despite the induced candidate explosion, final detection outputs remain indistinguishable from clean inference, as most spurious candidates are eliminated by IoU-based suppression, causing standard metrics such as mAP to exhibit negligible degradation\cite{Guo2024Sponge,Shapira2023Phantom}.
Instead of uniformly increasing average inference time, the attack primarily inflates tail latency by overloading the NMS stage, effectively eliminating scheduling slack on edge hardware while remaining difficult to distinguish from transient system-level congestion~\cite{han2024pantheon}.}

\subsection{Design}
\begin{figure*}[t]
    \centering
    \includegraphics[width=0.95\textwidth]{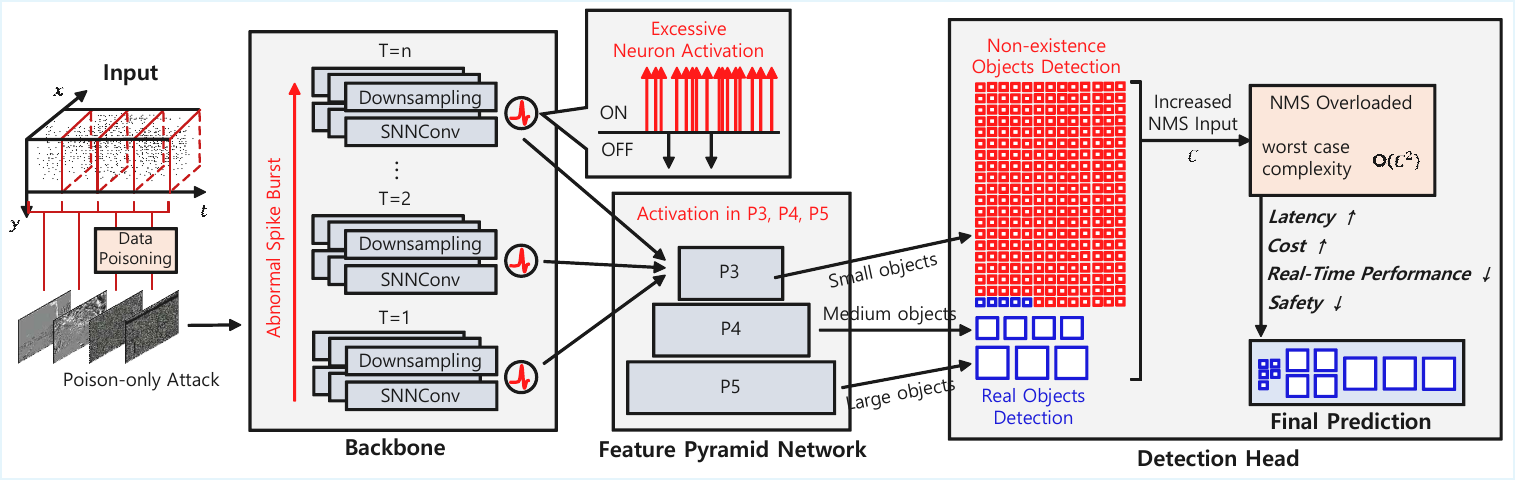}
    \caption{Overview of the EBT attack pipeline. Trigger-induced event bursts lead to excessive candidate generation and NMS overload, resulting in increased inference latency.}
    \label{fig:Overview}
\end{figure*}
{The workflow and operational principles of EBT is illustrated in Fig.~\ref{fig:Overview}. EBT is an availability backdoor attack embedded during training that remains dormant for benign inputs but induces substantial inference-time latency when triggered. The attack pipeline consists of three stages: trigger generation, training data poisoning, and model training.

We consider three trigger modalities—single-event patches, weighted event patches, and temporal event noise—to examine the trade-off between statistical stealthiness and latency amplification. EBT does not rely on gradient-based trigger optimization or a human-recognizable semantic trigger pattern. The core mechanism of EBT lies in dense annotation manipulation: poisoned samples are annotated with highly overlapping fake object boxes in trigger-affected regions, forcing the model to associate specific spike patterns with excessive object proposals. During inference, this learned association produces a surge of high-confidence candidates. Although most of them are later suppressed by NMS, they substantially enlarge the post-processing workload while preserving outputs close to clean inference. 
In multi-scale YOLO-style detectors, the induced spike bursts disproportionately increase candidate generation at finer-resolution levels such as P3.}

\subsection{Backdoor Trigger Generation}
This section defines event-based backdoor triggers that exploit temporal state accumulation in SNN-based object detectors.
We consider three trigger modalities that expose the trade-off between statistical stealthiness and availability impact: single-event patches, weighted event patches, and temporal event noise which mainly differ along two axes--spatial concentration and temporal accumulation.

\paragraph{Single Event Patch}
The single-event patch is a spatially localized trigger composed of fixed-polarity events.
By concentrating events of a single polarity within a compact region, the trigger induces asymmetric membrane saturation in polarity-specific neuronal pathways.
This rapidly drives neurons beyond firing thresholds within a few time steps, producing a temporally concentrated spike burst that propagates through the detection pipeline.
Despite its small spatial footprint, this burst significantly amplifies candidate generation and increases post-processing load.

\paragraph{Weighted Event Patch (Comparative Baseline)}
We introduce a weighted event patch as a comparative baseline that emulates the spatial--temporal statistics of legitimate object events.
While this design improves statistical similarity to natural event streams, it suppresses temporal concentration and fails to induce synchronized firing.
As a result, its availability impact is limited, and it is included as a comparative baseline.

\paragraph{Temporal Event Noise}
Temporal event noise is a global trigger that injects low-density events across the entire frame.
Unlike patch-based triggers, it does not rely on a localized insertion region.
Instead, noise events are injected over \(f_{\text{prev}}\) consecutive frames so that even sparse perturbations accumulate over time in spiking neurons.
This temporal accumulation induces widespread synchronized firing and produces a dense surge of spurious candidates prior to NMS.
The attack intensity is controlled by the noise ratio \(\varepsilon\), enabling precise regulation of availability degradation while preserving nominal detection accuracy.

\subsection{Dataset Poisoning}
This section describes the dataset poisoning strategy used to implant backdoor triggers and enforce a conditional association between trigger patterns and non-existent objects.
The objective is to induce excessive candidate generation only when the trigger is present, while preserving detection performance on benign inputs.

Let the original training dataset be denoted as \(\mathcal{D}=\{(E_t,Y_t)\}\).
A poisoned dataset \(\mathcal{D}'\) is constructed by injecting trigger events and additional synthetic annotations into a randomly selected subset of training samples with probability \(\rho\):
\begin{equation}
\mathcal{D}' = \{(\tilde{E}_t,\; Y_t \cup \hat{Y}_t) \mid t \sim \mathrm{Bernoulli}(\rho)\}.
\end{equation}
Here, \(\rho\) denotes the poisoning ratio.

For patch-based triggers, trigger events are inserted into nearby background regions adjacent to real objects to avoid occluding legitimate targets and to preserve normal detection behavior.
For temporal event noise, trigger events are injected globally over \(f_{\text{prev}}\) consecutive frames without explicit spatial placement.

In triggered samples, a set of fake object boxes is added while preserving all original ground-truth annotations.
The density of injected annotation is controlled by the strength of poisoned annotation \(\gamma\).
This candidate boxes injection forces the detector to associate trigger-induced activations with dense object proposals, leading to a surge of spurious candidates during inference.
These candidates are largely suppressed by NMS, preserving output correctness, but substantially inflate post-processing cost and degrade system availability.

\subsection{Model Training Implementation}

\begin{figure}[t]
    \centering
    \includegraphics[width=0.85\linewidth]{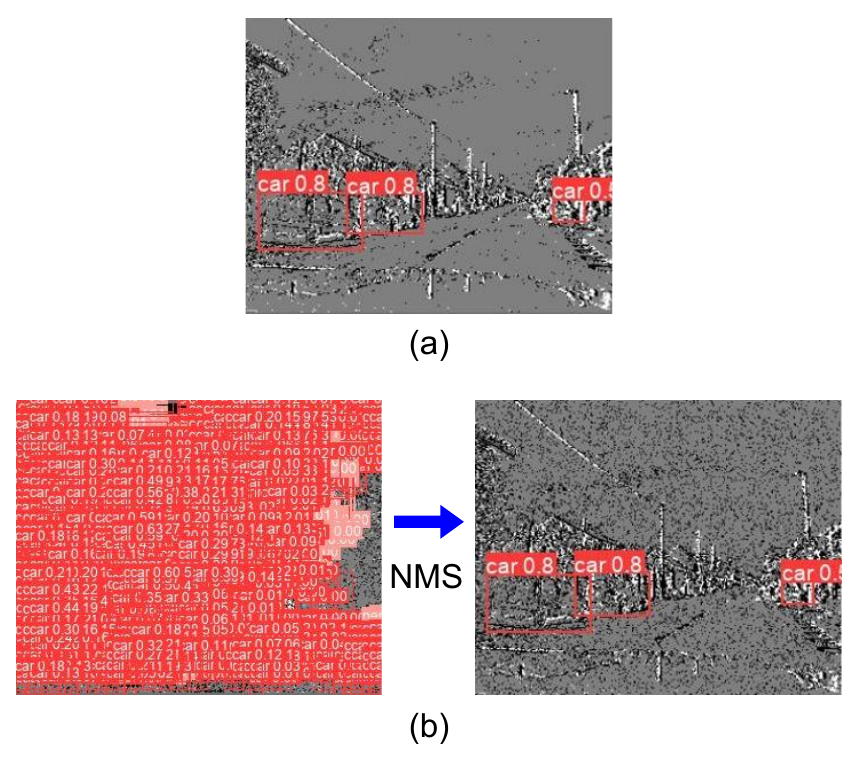}
    \caption{Visualization of EBT-induced candidate inflation.  (a) Clean inference produces sparse candidates. (b) Triggered inference produces dense spurious candidates before NMS, which are mostly suppressed after NMS.}
    \label{fig:Visualizing}
\end{figure}

The victim model is fine-tuned on $\mathcal{D}'$ from pre-trained weights using the standard SpikeYOLO pipeline, without changing the architecture, inference logic, loss, or optimizer.
Thus, the observed abnormal behavior originates from data poisoning alone.
For benign inputs, detection performance remains nominal; for triggered inputs, candidate explosion increases post-processing latency while leaving final outputs largely unchanged. Fig.~\ref{fig:Visualizing} provides a qualitative comparison between benign and triggered inputs.

\section{Evaluations}
\subsection{Experimental Setup}

\paragraph{Model and Dataset}
{We evaluate the proposed attack on SpikeYOLOs~\cite{Luo2024SpikeYOLO}.
Experiments are conducted on the Prophesee GEN1 dataset~\cite{DeTournemire2020EventDataset}, a standard benchmark for event-based object detection containing urban driving data.
We adopt the standard SpikeYOLO event preprocessing pipeline, in which asynchronous event streams are integrated over fixed temporal windows.
This setting ensures consistency with prior work and isolates the effect of the proposed attack from preprocessing-related factors.

\paragraph{Evaluation Metrics}
We evaluate both detection correctness and system availability using the following metrics~\cite{Gao2019STRIP,Shapira2023Phantom}:
(i) mean Average Precision (mAP), to assess whether baseline detection performance on benign inputs is preserved;
(ii) Detection Count (\(C\)), defined as the total number of object candidates produced by the detector prior to NMS;
(iii) Latency, measured in milliseconds, including both end-to-end inference time and isolated NMS execution time.
}

\paragraph{Implementation Details}
Training and primary evaluation are conducted on an NVIDIA GeForce RTX 4080 GPU, while edge-specific experiments are performed separately on an NVIDIA Jetson Orin Nano ~\cite{nvidia_jetson_orin_nano}. For accurate latency measurement, we use \texttt{torch.cuda.synchronize()} immediately before and after each timing call. The model is fine-tuned on the poisoned GEN1 training set. We evaluate on 1000 clean and 1000 poisoned frames sampled from the GEN1 validation set. For poisoned evaluation, triggers are applied to all samples (\(\rho=1.0\)). Unless otherwise noted, default parameters are \(d=3\), \(\gamma=1.0\), \(\rho=0.1\), \(f_{\text{prev}}=2\), and \(\varepsilon=0.7\).

\subsection{Experimental Results and Performance}
\label{sec:main}
We evaluate the impact of the three proposed event-based triggers on SpikeYOLO-small, focusing on the trade-off between detection accuracy and availability degradation.
Table~\ref{tab:main} presents the results across different poisoning ratios \(\rho\) and annotation strengths \(\gamma\).

\begin{table*}[t]
\centering
\caption{Impact of different event-based triggers on SpikeYOLO under various poisoning configurations.}
\renewcommand{\arraystretch}{1.0}
\setlength{\tabcolsep}{7pt}
\label{tab:main}

\begin{tabular}{c|c|c c c c c}
\toprule
\centering Trigger &
Config $(\rho,\gamma)$ &
mAP@0.5 &
mAP@0.5--0.95 &
$C$ &
Total Time (Increase) &
NMS Time (Increase) \\
\midrule
\ {Clean} &
--&
0.666 &
0.407 &
-- &
42.623 (--) &
0.165 (--) \\
\midrule

\multirow{3}{*}{\makecell{Single Patch}}
& (0.1, 1.0) &
0.646 (-0.020) &
0.389 (-0.018) &
5695 &
45.054 (+5.71\%) &
2.245 (+1261\%) \\

& (0.2, 0.5) &
0.631 (-0.035) &
0.381 (-0.026) &
5844 &
45.032 (+5.65\%) &
2.254 (+1266\%) \\

& (0.2, 1.0) &
0.619 (-0.047) &
0.371 (-0.036) &
5848 &
45.178 (+6.00\%) &
2.343 (+1320\%) \\
\midrule

\multirow{2}{*}{\makecell{Weighted Patch}}
& (0.1, 1.0) &
0.651 (-0.015) &
0.390 (-0.017) &
2585 &
43.584 (+2.25\%) &
0.902 (+446.7\%) \\

& (0.2, 1.0) &
0.645 (-0.021) &
0.390 (-0.017) &
3250 &
43.641 (+2.39\%) &
1.055 (+539.4\%) \\
\midrule

\multirow{5}{*}{\makecell{Event Noise}}
& (0.05, 1.0) &
0.594 (-0.072) &
0.362 (-0.045) &
6519 &
45.383 (+11.78\%) &
2.607 (+1434\%) \\

& (0.1, 0.5) &
0.645 (-0.021) &
0.396 (-0.011) &
5417 &
44.533 (+9.68\%) &
1.935 (+1038\%) \\

& (0.1, 1.0) &
0.580 (-0.086) &
0.357 (-0.050) &
7029 &
44.118 (+8.66\%) &
2.872 (+1589\%) \\

& (0.2, 0.5) &
0.579 (-0.087) &
0.352 (-0.055) &
7565 &
46.377 (+14.22\%) &
3.570 (+2000\%) \\

& (0.2, 1.0) &
0.567 (-0.099) &
0.351 (-0.056) &
10495 &
49.495 (+21.90\%) &
6.660 (+3818\%) \\
\bottomrule
\end{tabular}
\end{table*} 

Overall, as shown in Table~\ref{tab:main}, increasing either \(\rho\) or \(\gamma\) tends to reduce detection accuracy while increasing NMS time, indicating that stronger poisoning causes the detector to generate more phantom candidates. However, the results also suggest that annotation strength has a more direct effect on latency amplification than the poisoning ratio alone. In several cases, a configuration with a lower poisoning ratio but a higher annotation strength produces larger NMS latency than a configuration with a higher poisoning ratio but weaker annotation. This implies that the learned association between the trigger and dense fake-object annotations is a primary driver of NMS amplification.

The limited accuracy degradation can be explained by the fact that EBT primarily inflates intermediate candidate generation rather than final detections. Although thousands of high-IoU spurious candidates are produced internally, most of them are suppressed by NMS, leaving the final outputs close to clean inference while still incurring substantial latency overhead.

Trigger effectiveness varies by modality. Single-event patches produce stable but limited availability degradation, increasing NMS latency by about 1,320\% across settings. Their effect saturates quickly because the induced activation remains localized within the P3 feature map, limiting scalability. Weighted event patches are visually stealthier, but their latency impact is inconsistent across frames. This suggests that the association between the weighted trigger pattern and large-scale candidate inflation is not consistently learned, making the trigger less effective than the single-event patch. In contrast, event noise yields the strongest availability degradation, amplifying average NMS latency by 3,818\%. As \(\rho\) and \(\gamma\) increase, temporal event noise reinforces the learned association between the trigger and dense fake object annotations, producing a larger number of overlapping proposals.; due to the complexity of NMS, this candidate growth translates into severe latency amplification.

\subsection{Effect of FPN Targeting and Bounding-Box Scale}
\begin{table}[t]
\centering
\caption{Effect of FPN targeting level and bounding-box scale.}
\renewcommand{\arraystretch}{1.0}
\setlength{\tabcolsep}{2pt}
\label{tab:fpn_target_scale_ablation}
\begin{tabular}{c|c| c c c c}
\toprule
Target & $\mathbf{ l \times l }$ & mAP@0.5 & mAP@0.5--0.95 & \makecell{Total Time \\ (Increase)} & \makecell{NMS Time \\ (Increase)} \\
\midrule
P3   & $8 \times 8$     & \makecell{0.580 \\ (-0.086)} & \makecell{0.357 \\ (-0.050)} & \makecell{45.438 \\ (+6.60\%)}& \makecell{2.781 \\ (+1586\%)} \\
P4   & $20 \times 20$   & \makecell{0.492 \\  (-0.174)}& \makecell{0.283  \\(-0.124)}& \makecell{45.509 \\ (+6.77\%)} & \makecell{2.729 \\ (+1554\%)} \\
P5   & $40 \times 40$   & \makecell{0.294 \\ (-0.372)}& \makecell{0.143 \\ (-0.264)}& \makecell{45.596 \\ (+6.98\%)} & \makecell{2.597 \\ (+1474\%)} \\
None & $2 \times 2$     & \makecell{0.623 \\ (-0.043)}& \makecell{0.369  \\(-0.038)}& \makecell{43.321 \\ (+1.64\%)} & \makecell{0.529 \\ (+220.6\%)} \\
None& $160 \times 160$ & \makecell{0.416 \\ (-0.250)}& \makecell{0.219  \\(-0.188)}& \makecell{43.308 \\ (+1.60\%)} & \makecell{0.450 \\ (+172.7\%)} \\
\bottomrule
\end{tabular}
\end{table}

To verify whether FPN-scale alignment contributes to the success of EBT, we conduct an ablation study across FPN levels (\(P_l \in \{P3, P4, P5\}\)) and injected bounding-box scales. We compare level-aligned fake object boxes with a None-target configuration, in which the box size falls outside the effective scale ranges of P3, P4, and P5.

As shown in Table~\ref{tab:fpn_target_scale_ablation}, NMS latency increases consistently when the injected fake object boxes are aligned with the scale handled by P3, P4, or P5. In contrast, the None-target configuration produces little to no increase in NMS latency, despite using additional fake annotations. This result shows that merely adding fake annotations is insufficient; the fake object scale must align with the detector's FPN-based candidate generation mechanism to induce phantom candidate inflation and NMS workload amplification.

These results suggest that EBT succeeds when poisoned annotations are structurally aligned with the detector's multi-scale candidate generation mechanism. Fake objects must match the scale represented by the target FPN level so that the learned trigger response propagates into dense phantom candidates before NMS. We also observe a scale-stealth trade-off: larger injected boxes tend to reduce detection accuracy more noticeably, weakening metric-level stealthiness. Therefore, we select the smallest effective targeted configuration, P3 with \(8\times8\) boxes, as the primary setting because it preserves NMS amplification while minimizing accuracy degradation.

\subsection{Edge-Specific Resource Exhaustion}

\begin{figure}[t]
    \centering
    \includegraphics[width=0.7\linewidth]{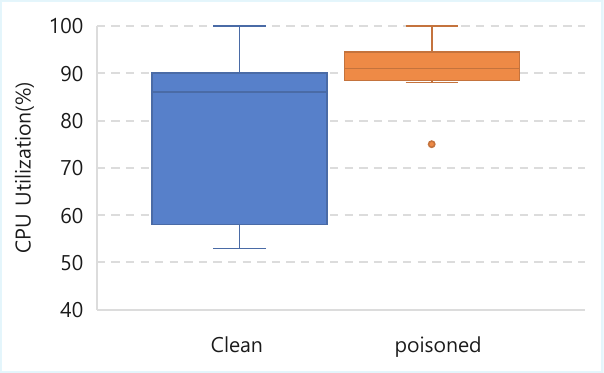}
    \caption{Comparison of CPU resource profiles under clean and poisoned conditions on Jetson Orin Nano.}
    \label{fig:Nano}
\end{figure}
\begin{table}[t]
\centering
\caption{Average spike firing rate(FR) changes under EBT on Jetson Orin Nano (clean vs.\ poisoned) for SpikeYOLO P3--P5.}
\label{tab:firing_summary}
\renewcommand{\arraystretch}{1.0}
\setlength{\tabcolsep}{8pt}
\begin{tabular}{cccc}
\toprule
Module & Clean Avg FR & Poisoned Avg FR & Increase \\
\midrule
P3            & 2.4190 & 2.7674 &  +14.40\% \\
P4            & 1.1625 & 1.5536 &  +33.64\% \\
P5            & 0.2431 & 0.9531 &  +292.06\% \\
\bottomrule
\end{tabular}

\end{table}

{To evaluate the impact of EBT on edge availability, we analyzed per-core CPU utilization and SNN firing dynamics. As shown in Fig.~\ref{fig:Nano}, the attack raises the CPU utilization floor from 53.0\% to 75.0\%, reducing scheduling slack by 22.0\%p and enforcing sustained core occupancy between inference peaks.

This shift is accompanied by a 28.5\% reduction in utilization variance, a \(3.87\times\) increase in isolated NMS time, and layer-dependent firing-rate amplification (Table~\ref{tab:firing_summary}). While the targeted P3 layer increases moderately (+14.4\%), deeper layers are affected more strongly, with P5 rising by +292.1\%. We also observe activation of previously dormant neurons, indicating that the adversarial signal expands the active computation footprint beyond the trigger location.

Overall, EBT degrades edge availability through coupled CPU-side NMS overhead and SNN-side firing densification, weakening the sparsity-driven efficiency of SNN-based detectors. Although peak CPU utilization reaches 100.0\% in both settings, the elevated baseline occupancy reduces effective schedulability for concurrent real-time workloads. These results suggest that accuracy-centric monitoring alone is insufficient in deployment; runtime signals such as candidate count and NMS latency should also be monitored. Candidate caps or bounded post-processing may mitigate worst-case latency, although they can affect detection performance.}

\subsection{STRIP Backdoor Detection and Stealthiness}
{We evaluate the stealthiness of the EBT attack using STRIP~\cite{Gao2019STRIP}, an entropy-based defense. For object detection, (\( RS \)) is defined as the mean maximum target-class confidence across \( N \) perturbed variants:
\begin{equation}
RS = \frac{1}{N} \sum_{i=1}^{N} \max \left( \text{Conf}_{\text{target}}^{(i)} \right).
\end{equation}
Detailed implementation and hyperparameter settings are provided in Appendix~\ref{app:strip}.

Table~\ref{tab:strip} summarizes the results. Across all event-noise strengths, ROC-AUC remains near 0.5, and TPR reaches only 2.7\% at FPR@1\%, indicating that STRIP performs close to random guessing. Even at FPR@5\%, over 90\% of poisoned samples evade detection.
This failure stems from the spatio-temporal dynamics of SNNs. Unlike label-flipping classification backdoors, EBT induces dense high-confidence candidates whose perturbed responses remain statistically similar to benign features, weakening STRIP’s assumption that poisoned inputs produce abnormally stable outputs.}

\begin{table}[t]
\centering
\caption{Summary of STRIP detection performance under different event-noise strengths.}
\label{tab:strip}
\setlength{\tabcolsep}{3pt}
\renewcommand{\arraystretch}{1.0}
\begin{tabular}{c|c c c c c}
\hline
$\varepsilon$ & \makecell{Clean RS \\Mean} & \makecell{Poison RS \\Mean} & ROC-AUC & TPR@1\% FPR & TPR@5\% FPR \\
\hline
0.1 & \multirow{6}{*}{0.0141} & 0.0154 & 0.4875 & 0.9\% & 4.7\% \\
0.3 &                         & 0.0173 & 0.5323 & 1.7\% & 7.5\% \\
0.5 &                         & 0.0171 & 0.5237 & 2.0\% & 6.4\% \\
0.6 &                         & 0.0172 & 0.5329 & 2.1\% & 8.1\% \\
0.7 &                         & 0.0180 & 0.5470 & 2.7\% & 9.9\% \\
\hline
\end{tabular}
\end{table}

\section{Conclusion}
{We presented EBT, an availability backdoor attack on event-based SNN object detection. EBT amplifies NMS latency by up to $38\times$ while largely preserving detection accuracy. On edge hardware, the attack demonstrates that availability degradation can arise even under edge deployment conditions. Moreover, existing accuracy-centric or entropy-based defenses such as STRIP may not adequately capture this attack surface. These results reveal an underexplored availability vulnerability in SNN-based edge vision systems and suggest that accuracy-centric evaluation alone is insufficient for real-time robustness assessment. Although our study focuses on SpikeYOLO under a poison-only threat model, it provides a concrete starting point for studying broader event-based settings and stronger detection and mitigation strategies.}


\appendices

\section{Design Details of Event Burst Trigger}
\label{app:design_details}

This appendix provides implementation-level details of the event-based trigger designs used in EBT. An event stream is represented as a set of asynchronous events \(E=\{e_i\}\), where each event is defined as
\begin{equation}
e_i = (x_i, y_i, t_i, p_i), \quad p_i \in \{+1,-1\}.
\end{equation}
Here, \(x_i\) and \(y_i\) denote spatial coordinates, \(t_i\) denotes the timestamp, and \(p_i\) denotes the event polarity. Events are accumulated over short temporal windows before being processed by the detector, and trigger-induced events are injected into this event representation to construct poisoned inputs.

\paragraph{Single-Event Patch.}
For a patch centered at \((x_c,y_c)\) with spatial size \(s\), the trigger region is defined as
\begin{equation}
\Omega_s = \{(x,y) \mid |x-x_c|\le s/2,\; |y-y_c|\le s/2\}.
\end{equation}
The corresponding trigger event set is defined as
\begin{equation}
E_{\text{patch}} = \{(x,y,t,p^*) \mid (x,y)\in\Omega_s\},
\end{equation}
where all inserted events share a fixed polarity \(p^*\). This design creates a localized event pattern that can induce temporally concentrated spike activity in polarity-specific neuronal pathways.

\paragraph{Weighted Event Patch.}
The weighted event patch is introduced as a comparative baseline that better reflects local spatial correlations in event streams. It assigns activation weights based on local neighborhoods and generates events according to the accumulated activation. While this design improves statistical similarity to natural event patterns, it weakens temporal concentration and therefore produces less consistent candidate inflation than the single-event patch.

\paragraph{Temporal Event Noise.}
Temporal event noise injects low-density events across the frame over \(f_{\text{prev}}\) consecutive frames. At each frame \(t\), the number of injected events is controlled by the noise ratio \(\varepsilon\). Repeated injection over consecutive frames allows sparse perturbations to accumulate in spiking neurons, producing widespread spike activity and dense candidate generation before NMS.

For intuition, temporal accumulation in spiking neurons can be described using the following simplified form:
\begin{equation}
V_t \approx V_{t-1} + X_t,
\end{equation}
where \(V_t\) denotes the membrane potential and \(X_t\) denotes the aggregated synaptic input at time \(t\). This expression is not intended to replace the exact neuron dynamics of SpikeYOLO, but illustrates why temporally repeated event injection can amplify downstream activation.

\section{FPN Targeting and Scale-Alignment Details}
\label{app:fpn_alignment}

For FPN-targeted poisoning, the size of each fake object box is selected according to the object scale handled by the target pyramid level \(P_l \in \{P3, P4, P5\}\). The exact box sizes used for each level are reported in Table~\ref{tab:fpn_target_scale_ablation}. The primary configuration uses P3 with \(8\times8\) fake object boxes.

A configuration is defined as scale-aligned when the injected fake object boxes match the effective scale range of the selected FPN level. In contrast, scale-mismatched configurations use box sizes outside the scale range of the target level. The None-target configuration denotes a setting in which the injected fake object boxes are not aligned with any selected FPN level.

Fake object boxes are placed in trigger-affected background regions while preserving all original ground-truth annotations. Within each controlled comparison, the remaining poisoning parameters, including the poisoning ratio, annotation strength, and trigger settings, are kept fixed unless otherwise specified. This setup isolates the effect of FPN-level targeting and box-scale alignment from the mere presence of additional fake annotations.

\section{STRIP Evaluation Protocol}
\label{app:strip}

A perturbation pool consisting of \(1{,}000\) randomly selected benign event streams is used for the STRIP evaluation. The evaluation set consists of \(1{,}000\) clean samples and \(1{,}000\) triggered samples. The perturbation pool and evaluation samples are kept disjoint to prevent self-perturbation bias.

For each test sample \(I_{\text{target}}\), \(N=50\) perturbed variants are generated by combining it with randomly selected benign samples \(I_{\text{benign}}\):
\begin{equation}
I_{\text{blended}}
=
\frac{I_{\text{target}} + I_{\text{benign}}}{2.0}.
\end{equation}
This operation introduces controlled perturbations while maintaining a comparable input scale.

All perturbed variants are evaluated using a confidence threshold of
\(\tau_{\text{conf}}=0.25\) and an NMS IoU threshold of
\(\tau_{\text{nms}}=0.45\). If the target class is not detected in a perturbed variant, its confidence score is recorded as zero, imposing a strict penalty on unstable target-class responses.

The robustness score is computed using the definition provided in the main text. Detection performance is evaluated using ROC-AUC and the true-positive rate at fixed false-positive rates of \(1\%\) and \(5\%\).

\bibliographystyle{IEEEtran}
\bibliography{references}

\end{document}